\documentclass[runningheads]{llncs}
\usepackage{graphicx}
\usepackage{algorithm}
\usepackage{algpseudocode}
\usepackage{amsmath}
\usepackage{amssymb}
\usepackage{booktabs}
\usepackage{subfig}

\begin{document}
\title{Vision Conformer: Incorporating Convolutions into Vision Transformer Layers\thanks{This research was partially supported by MEXT-Japan (Grant No. JP23K16949).}}
\titlerunning{Vision Conformer}
% \author{Anonymous}
\author{Brian Kenji Iwana\inst{1}\orcidID{0000-0002-5146-6818} \and
Akihiro Kusuda\inst{2}\thanks{Equal contribution}}
% %
\authorrunning{B. K. Iwana and A. Kusuda}
\institute{Kyushu University, Fukuoka, Japan \and
Nara Institute of Science and Technology, Nara, Japan \\
\email{iwana@ait.kyushu-u.ac.jp}}
\maketitle              % typeset the header of the contribution
\begin{abstract}
Transformers are popular neural network models that use layers of self-attention and fully-connected nodes with embedded tokens. Vision Transformers (ViT) adapt transformers for image recognition tasks. In order to do this, the images are split into patches and used as tokens. One issue with ViT is the lack of inductive bias toward image structures. Because ViT was adapted for image data from language modeling, the network does not explicitly handle issues such as local translations, pixel information, and information loss in the structures and features shared by multiple patches. Conversely, Convolutional Neural Networks (CNN) incorporate this information. Thus, in this paper, we propose the use of convolutional layers within ViT. Specifically, we propose a model called a Vision Conformer (ViC) which replaces the Multi-Layer Perceptron (MLP) in a ViT layer with a CNN. In addition, to use the CNN, we proposed to reconstruct the image data after the self-attention in a reverse embedding layer. Through the evaluation, we demonstrate that the proposed convolutions help improve the classification ability of ViT.     
\end{abstract}

\begin{figure}
\centering
\subfloat[Vision Transformer]{\includegraphics[width=0.4\linewidth]{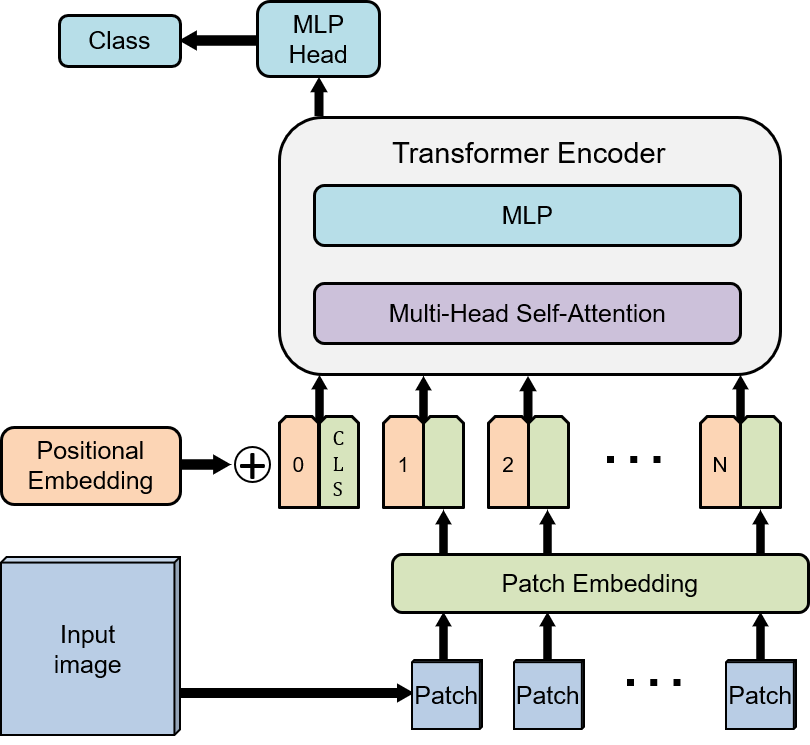}%
\label{subfig:vit}}
\hfil
\subfloat[Vision Conformer (Proposed)]{\includegraphics[width=0.4\linewidth]{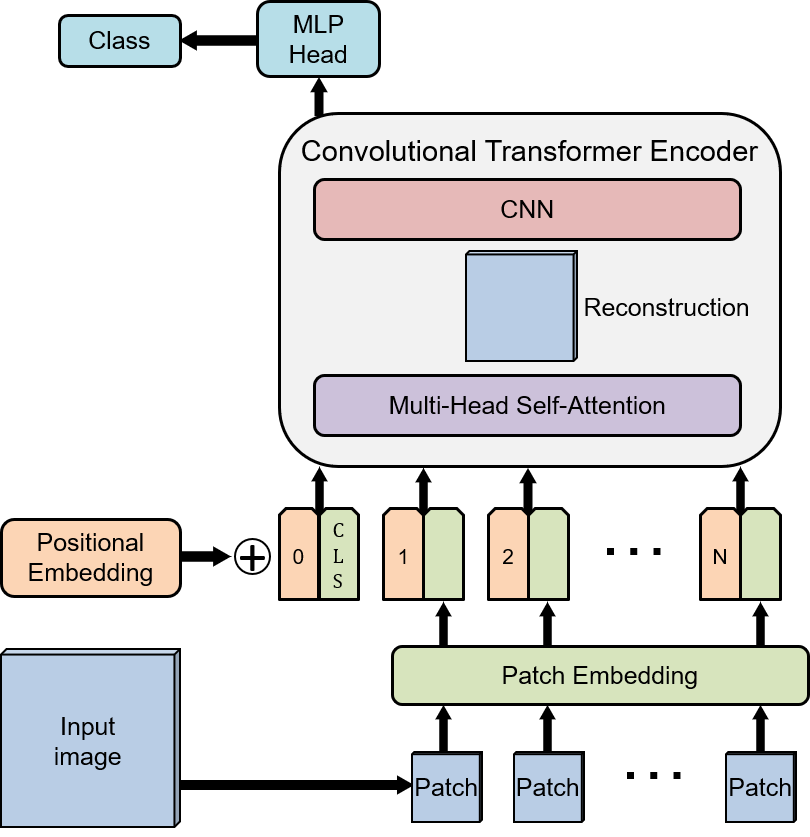}%
\label{subfig:vic}}
\caption{Comparison of the overall structure of ViT and the proposed method.}
\label{fig:vic}
\end{figure}

\section{Introduction}

% intro
Recently, there has been a sudden rise in the popularity of Transformers~\cite{vaswani2017attention} in pattern recognition. 
They were originally proposed and popularized in the Natural Language Processing~(NLP) domain~\cite{vaswani2017attention,devlin2018bert,radford2019language}. 
Fundamentally, Transformers are feed-forward neural networks that utilize self-attention, inner-layer residual connections, and a specialized token-based encoding. 
Typically, shown in Fig.~\ref{subfig:vit}, Transformers are designed with multiple Transformer blocks, each containing a Multi-Head Self-Attention~(MHSA) layer and a fully-connected layer, and are trained with an encoder and a decoder network. 
The self-attention layer uses an attention mechanism to relate the weighted positions of an input sequence with the other positions of the input. 
This self-attention helps the fully-connected nodes to put more emphasis on the pairwise relationships of the input.

% vision transformers
The popularity of Transformers has also spread to image recognition tasks. 
Part of the trend in image-based Transformers usage is due to the success of Transformers in image classification and object detection. 
For example, in seminal works, Dosovitskiy et al.~\cite{dosovitskiy2020image} proposed the Vision Transformer~(ViT) which breaks images into patches to be used as tokens, and Chen et al.~\cite{chen2020generative} and Parmar et al.~\cite{parmar2018image} use a sequence Transformer to do row-wise generation of images.
More recently, Transformer-based models hold the state-of-the-art results on many computer vision benchmarks, including ImageNet~\cite{dai2021coatnet,zhai2021scaling}, COCO~\cite{xu2021end,yuan2021florence}, CIFAR10/100~\cite{dosovitskiy2020image,touvron2021going,ridnik2021ml}, and more.

%ViT
Introduced by ViT, a common mechanism in image-based Transformers is to break the input image into fixed-sized patches and embed the patches that are linearly transformed into 1D vectors. 
This is done to adapt the images to be used with traditionally sequential word piece token representations. 
The subsequent patch-based token representations are used by the self-attention and fully-connected layers of the Transformer. 

% % problems 
One problem with ViT is that it is possible for the information to be lost in the fixed-sized patch-based representation.
Since the patches are arbitrarily divided, there is no consideration for the objects, features, or structures that could exist across or between multiple patches. 
This is because the self-attention mechanism and fully connected layers only consider the patches as a whole. 
It should be noted, that one solution to this problem is the use of hierarchical transformers that use different-sized patches~\cite{Liu_2021}.
In addition, while self-attention does create global relationships, the relationships are only pairwise relationships and can only rely on the positional embedding for global structure. 

\begin{figure}
\centering
\includegraphics[width=1\linewidth]{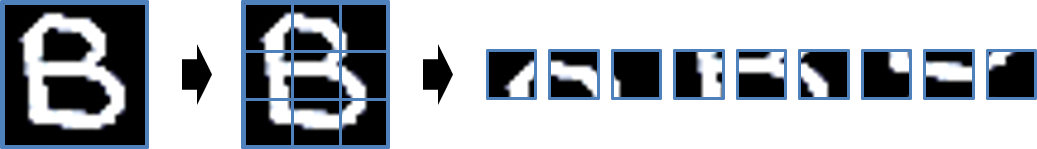}%
\caption{An example of how ViT turns an image of a character into a sequence of patches.}
\label{fig:character}
\end{figure}

% font recognition
This is especially true for character recognition. 
Due to patches being fixed size and location, small translations in the location of the character can have a large effect on the patch embedding.  
Furthermore, as shown in Fig.~\ref{fig:character}, in character recognition, the character occupies multiple patches, and consideration for cross-patch structures is not performed.

% convolutions
Conversely, Convolutional Neural Networks~(CNN)~\cite{Lecun_1998} have been shown to overcome these issues. 
The use of convolutional layers and max pooling allows for some translation invariance~\cite{Lecun_1998}.
Unlike fixed-size patch-based Transformers, the embedding for the fully-connected layers in a CNN can represent large overlapping receptive fields. 

%solution
CNNs have shown to be a powerful tool in computer vision and pattern recognition~\cite{Schmidhuber_2015}. 
Thus, there is a desire to combine the advantages of CNNs and ViT. 
Consequentially, many convolutional variations of ViT have been proposed. 
Many Transformers, such as Convolutional vision Transformer~(CvT)~\cite{wu2021cvt}, Compact Convolutional Transformer~(CCT)~\cite{hassani2021escaping}, and Convolution and Self-Attention Network~(CoAtNet)~\cite{dai2021coatnet} realize the advantages of feature extraction using convolutional layers. 
Therefore, they combine CNNs with ViT by using convolutional layers to improve the embeddings for Transformers.

\begin{figure}
\centering
\subfloat[Vision Transformer]{\includegraphics[width=0.6\linewidth]{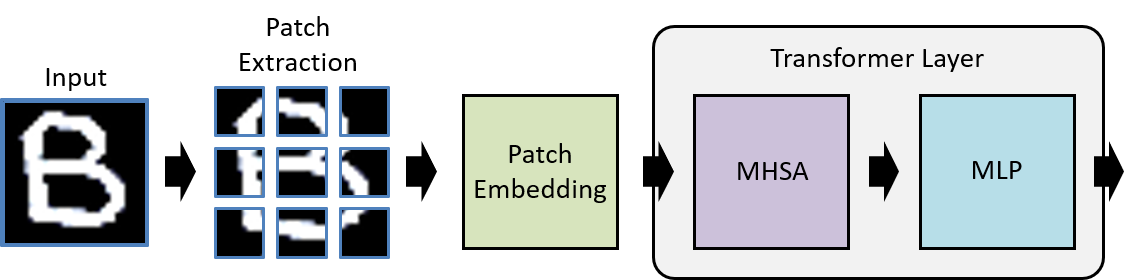}%
\label{subfig:vitprocess}}
\hfil
\subfloat[Vision Conformer (Proposed)]{\includegraphics[width=1\linewidth]{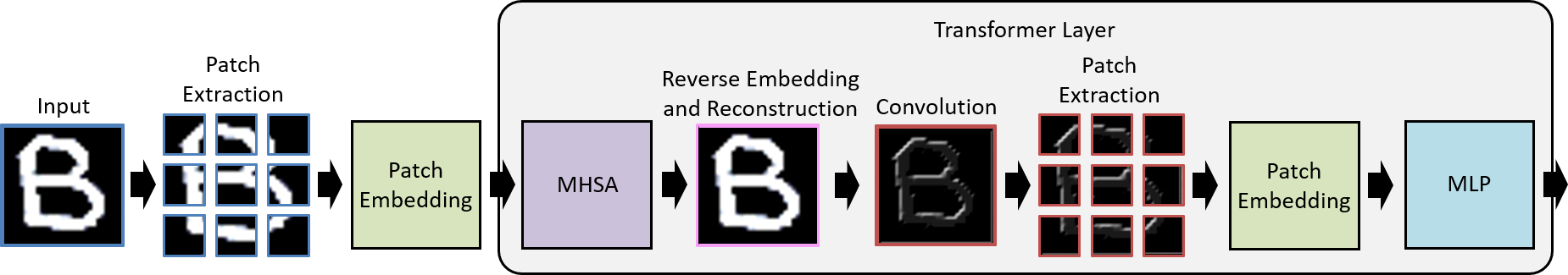}%
\label{subfig:vicprocess}}
\hfil
\subfloat[Other Convolutional Transformers]{\includegraphics[width=0.75\linewidth]{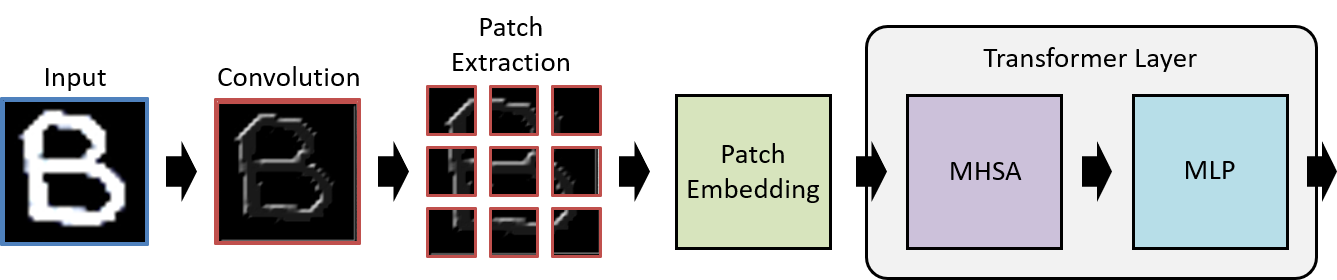}%
\label{subfig:convprocess}}
\caption{Comparison of the steps taken in ViT, ViC, and other convolutional transformers like CCT and CvT.}
\label{fig:process}
\end{figure}

In order to take advantage of the feature extraction, slight translation invariance, and structure-preserving abilities of CNNs, we propose to incorporate a convolutional layer inside the Transformer block. 
As shown in Fig.~\ref{fig:process}, the spatial outputs of the self-attention layer are reconstructed into an image and then provided to convolutional layers. 
After the convolutional layers, a fully-connected layer is used, much like a typical Transformer. 
Another interpretation of this would be replacing the Multi-Layer Perceptron~(MLP) of ViT with a CNN. 

% contributions
The contributions of this paper are as follows:
\begin{itemize}
    \item We propose a new convolutional vision Transformer called a Vision Conformer~(ViC). The proposed ViC replaces the MLP of ViT with a CNN.
    \item In order to adapt the Transformer block to use a convolution, we introduce a reverse embedding layer to reconstruct the patches from the vector embeddings and a reconstruction module to recompose the patches into a matrix for the convolutional layers. 
    \item The proposed method is evaluated on the character recognition task. We show results on three common character recognition datasets, MNIST, EMNIST, and KMNIST. We compare the proposed method to five state-of-the-art convolutional transformers and ViT.
\end{itemize}

The code is publicly available at \url{https://github.com/uchidalab/vision-conformer}.

\section{Related Work}

The most common way to use convolutions with transformer or self-attention-based models is to use convolutional layers to extract features for the embeddings. 
For example, Coccomini et al.~\cite{coccomini2021combining} used EfficientNet~\cite{tan2019efficientnet} to extract features for the patches of ViT and Cross-Attention Multi-Scale Vision Transformer (CrossViT)~\cite{chen2021crossvit}. 
CvT~\cite{wu2021cvt} uses convolutions to generate the token embeddings. 
Similarly, Convolution-enhanced image Transformers~(CeiT)~\cite{yuan2021incorporating} use depth-wise convolutions to locally enhance the tokens of a ViT.
Hassani et al. proposed the CCT~\cite{hassani2021escaping} that replaces the standard ViT patch embedding layer with a convolutional layer.
CoAtNet~\cite{dai2021coatnet} uses a hybrid of a CNN and Transformer by including convolutional layers for the lower layers and Transformer blocks for the upper layers.
In another work, Chu et al.~\cite{chu2021conditional} propose a modified positional embedding for ViT called Conditional Positional Embeddings~(CPE). 
In their proposed Conditional Positional encoding Vision Transformer~(CPVT), the CPEs are constructed using convolutions. 
Pooling-based Vision Transformer~(PiT)~\cite{heo2021rethinking} uses convolutions with stride 2 to downsample the spatial embeddings of ViT. 

In addition, there are other ways convolutions are used in ViT-based models. 
For example, Convolutional Vision Transformer~(ConViT)~\cite{d2021convit} proposes Gated Positional Self-Attention~(GPSA) that can be initialized as a convolutional layer. 
Through this, the GPSA can learn similar properties as a convolution. 
In Zhang et al.~\cite{zhang2022nested}, a convolution is used to aggregate hierarchical transformer blocks in their Nested Hierarchical Transformer~(NesT).

There are also CNN models that utilize self-attention. 
For example, CNNs using self-attention have been used for medical imaging~\cite{Wu_2019,Li_2020}. The self-attention helps establish relationships between regions of the image that is beyond the receptive field of the convolutions.
CNN Meets Transformer~(CMT)~\cite{guo2021cmt} uses alternating convolutional blocks and lightweight self-attention blocks with local perception units. 
Transformer in Convolutional Neural Networks~(TransCNN)~\cite{liu2021transformer} utilize hierarchical MHSA~(H-MHSA) blocks within a CNN.

Compared to these methods, as far as the authors know, the proposed method is the only method to incorporate the convolutional layers directly into the transformer layers of a transformer model. 
Most methods that incorporate convolutions into their architectures usually either use the convolutions separate from the transformer layers or just use the self-attention as part of a CNN.

In time series recognition, namely speech recognition, the Conformer~\cite{Gulati_2020} is a transformer that contains a convolution between the self-attention layer and the fully-connected layer. 
Due to the similarity of the Conformer and the proposed method, we adopt a similar name. 
However, the proposed method was developed without inspiration or relation to a Conformer. 
And unlike the Conformer, the proposed method is based on ViT and requires extra consideration for the patch embeddings and image-based convolutions. 

\section{Vision Transformers (ViT)}

% \subsection{Transformers}

Transformers are neural network models that were originally designed for NLP~\cite{vaswani2017attention}. 
They are constructed from Transformer blocks consisting of a self-attention layer followed by a fully-connected layer with residual connections between layers. 
In the traditional NLP Transformers, text is modeled using sequences of word part tokens. 
These tokens are embedded into vectors and fed to the Transformer.

ViT~\cite{dosovitskiy2020image} is a Transformer that was adapted for image recognition. 
The novel idea of ViT is that they proposed to use small image patches from the input images as tokens instead of the traditional word part tokens. 
Another difference is the traditional Transformer is trained in an encoder-decoder structure. 
Comparatively, ViT only uses an encoder during training.
In this section, we will provide background and describe the important features of ViT.

\subsection{Image Tokenization}

In order to adapt images to be used as sequences of token embeddings, ViT breaks the image into fixed-sized patches, as shown in Fig.~\ref{fig:vic}. 
The patches are serialized into a sequence. 
Each element of the sequence is flattened and embedded into a vector using a trainable linear projection. 
The result is a sequence $\mathbf{X}=\mathbf{x}_1,\dots,\mathbf{x}_t,\dots,\mathbf{x}_T$, where $\mathbf{x}_t$ is a vector embedding of each patch and $T$ is the number of patches. 
Because the patches are serialized, structures that span multiple patches are arbitrarily split based on the patch size.

Furthermore, a 1D positional embedding is added to the patch embedding.
The positional embedding indicates the position of the token in the sequence. 
The purpose of the positional embedding is to retain the positional information of the patches. 

\subsection{Classification Tokenization}

In addition to the patch embedding a special token is used to indicate the embedding to use for the classifier. 
This special classification token is prepended as $\mathbf{x}_{\mathrm{CLS}}$ to $\mathbf{X}$ as the first element of the sequence, i.e. $\mathbf{X}=\mathbf{x}_\mathrm{CLS},\mathbf{x}_1,\dots,\mathbf{x}_t,\dots,\mathbf{x}_T$. 
In each layer of ViT, the first element of the sequence of embeddings is the classification token and the subsequent elements are the patch tokens. 
For classification, an MLP head is used as a classifier by attaching a fully-connected layer to the output embedding of the topmost classification token.

\subsection{Multi-Head Self-Attention}

Self-attention is a special case of attention where elements of the input sequence are weighted and multiplied with themselves. 
Specifically, the input sequences are copied into a query $\mathbf{Q}$, key $\mathbf{K}$, and value $\mathbf{V}$ and weighted separately. 
The elements of query $\mathbf{Q}$ and key $\mathbf{K}$ are multiplied and become the attention mechanism for value $\mathbf{V}$.
In Transformers, Scaled Dot-Product Attention is used for the self-attention mechanism. 
Namely, the Scaled Dot-Product Attention is defined as:
\begin{equation}
    \mathrm{Attention}(\mathbf{Q},\mathbf{K},\mathbf{V})=\mathrm{softmax}\left(\frac{\mathbf{Q}\mathbf{K}^T}{\sqrt{d}}\right)\mathbf{V}, 
\end{equation}
where $1/\sqrt{d}$ is a scaling factor by the number dimensions $d$ of the input sequences. 
The idea of self-attention is that the important pairwise relationships between tokens should be emphasized in the representation for the fully-connected layer.

%multi-head
Multi-head attention is attention that uses more than one parallel attention blocks. 
In Transformers, for each $\mathbf{Q}$, $\mathbf{K}$, and $\mathbf{V}$, multiple self-attention layers are used and the results are concatenated. 
This is done to jointly attend different combinations of pairwise matches simultaneously.

\subsection{Multi-Layer Perceptron (MLP)}

After the multi-head self-attention layer, ViT and other Transformers use a fully-connected MLP layer. 
The input to the MLP layer is the sequence output by self-attention, including the vector embedding related to the classification token. 
Also, between each layer Layer Normalization~\cite{ba2016layer} and a residual connection~\cite{He_2016} is used.

\section{Vision Conformer (ViC)}

The proposed Vision Conformer~(ViC) is modeled on ViT~\cite{dosovitskiy2020image}. 
As shown in Fig.~\ref{fig:vic}, we adopt a similar structure, including the same patch tokenization and embedding and self-attention. 
The difference between ViT and ViC is that we propose to replace the MLP of ViT with a CNN.

\begin{figure}
    \centering
    \includegraphics[width=0.5\linewidth]{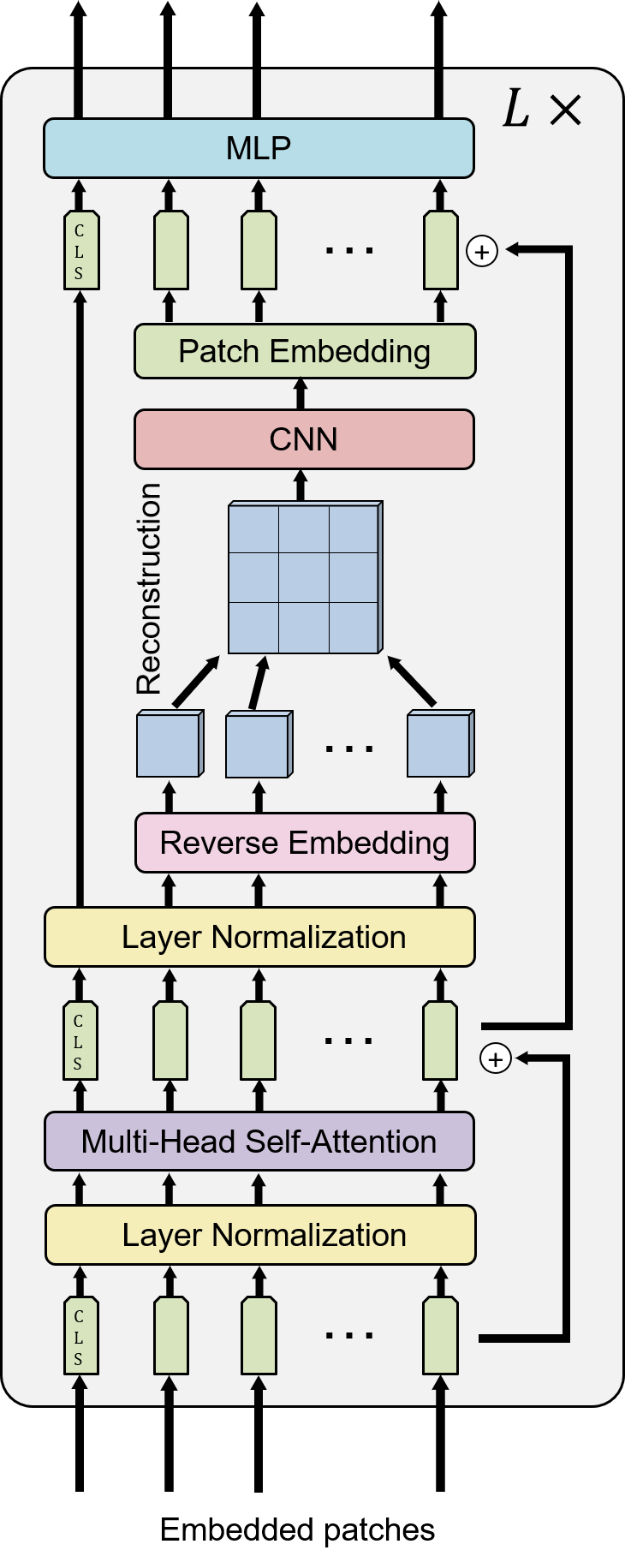}
    \caption{Details of a ViC Encoder block within the proposed ViC. The ViC Encoder is similar to a standard Transformer Encoder, except that after the Multi-Head Self-Attention layer, the patch embeddings are reconstructed into an image and a CNN is used instead of the standard MLP. To reconstruct the image, a Reverse Embedding layer is used. Note, special tokens such as the classification token are passed directly to the fully connected layer.}
    \label{fig:encoder}
\end{figure}

Fig.~\ref{fig:encoder} details the proposed ViC Encoder block. 
Similar to ViT, the embedded patches are input and Layer Normalization~\cite{ba2016layer} is applied. 
Next, MHSA is used. 
The residual connection is then added to the output of MHSA and Layer Normalization is applied again. 
After Layer Normalization, the hidden vector $\mathbf{Z}^\mathrm{(in)}=\mathbf{z}_\mathrm{CLS},\mathbf{z}_1,\dots,\mathbf{z}_t,\dots,\mathbf{z}_T$ is split into the classification embedding $\mathbf{z}_\mathrm{CLS}$ and the patch embeddings $\mathbf{z}_1,\dots,\mathbf{z}_t,\dots,\mathbf{z}_T$. 
This is done because the classification embedding $\mathbf{z}_\mathrm{CLS}$ is not part of the image representation.

From there, we introduce a Reverse Embedding step, a Reconstruction module, a CNN, and a Patch Embedding step that are applied to the patch embeddings $\mathbf{Z}_\mathrm{patch}$. 
Finally, the hidden embeddings are fed to the fully-connected MLP.

The input of this encoder block is the same patch embeddings with positional encodings as ViT. 
Furthermore, the block can be stacked $L$ number of times. 
To use the proposed method for classification, an MLP head is used on the final $\mathbf{z}_\mathrm{CLS}$ embedding.

\subsection{Image Reconstruction}

\begin{algorithm}[!t]
\caption{Reverse Embedding, Reconstruction, and CNN}\label{alg:vic}
\begin{algorithmic}[1]
\Require $\mathbf{Z}^\mathrm{(in)}=\mathbf{z}_\mathrm{CLS},\mathbf{z}_1,\dots,\mathbf{z}_t,\dots,\mathbf{z}_T$
\Ensure $\mathbf{Z}'^\mathrm{(out)}=\mathbf{z}'_\mathrm{CLS},\mathbf{z}'_1,\dots,\mathbf{z}'_t,\dots,\mathbf{z}'_T$
    \State $\mathbf{z}_\mathrm{CLS}, \mathbf{Z}_\mathrm{patch}\gets\mathbf{Z}^\mathrm{(in)}$ \Comment{Separate the patch embeddings}
    \State $\mathbf{Z}_\mathrm{patch}\gets\mathrm{Linear}(\mathbf{Z}_\mathrm{patch})$ \Comment{Reverse embedding}
    \State $\mathbf{P}\gets\mathrm{Reshape}(\mathbf{Z}_\mathrm{patch})$ 
    \State $\mathbf{I}\gets\mathrm{SpatialConcatenate}(\mathbf{P})$ \Comment{Reconstruction}
    \For{$c\gets1,C$} \Comment{CNN}
    \State $\mathbf{I}'\gets\mathrm{Convolution}(\mathbf{I})$
    \EndFor
    \State $\mathbf{P}'\gets\mathrm{PatchExtraction}(\mathbf{I}')$
    \State $\mathbf{Z}'_\mathrm{patch}\gets\mathrm{Flatten}(\mathbf{P}')$ \Comment{Patch embedding}
    \State $\mathbf{Z}'_\mathrm{patch}\gets\mathrm{Linear}(\mathbf{Z}'_\mathrm{patch})$
    \State $\mathbf{Z}'\gets\mathrm{Concatenate}(\mathbf{z}_\mathrm{CLS}, \mathbf{Z}'_\mathrm{patch})+\mathbf{Z}^\mathrm{(in)}$
    \State $\mathbf{Z}'^\mathrm{(out)}=\mathrm{MLP}(\mathbf{Z}')$
    \end{algorithmic}
\end{algorithm}

In order to use a 2D CNN within a Transformer block, the output of MHSA needs to be a matrix. 
Thus, we reconstruct the image structure from the patch embeddings. 
To do this, we introduce two techniques, \textit{Reverse Embedding} and \textit{Reconstruction}. 
The process of Reverse Embedding and Reconstruction is the reverse operation of patch embedding used on the input of ViC. 
Specifically, the Reverse Embedding returns the latent vectors back to the original shape of the patches and the Reconstruction module recombines the patches into a single matrix. 

Algorithm~\ref{alg:vic} shows the steps required to reconstruct the image from the token embeddings. 
In Algorithm~\ref{alg:vic}, the input is the token embeddings $\mathbf{Z}^\mathrm{(in)}$ from the output of MHSA. 
The output $\mathbf{Z}'^\mathrm{(out)}$ is either passed to the next ViC layer or is provided to the MLP head for classification. 
$\mathbf{P}$ is the reconstructed patches from the patch embeddings, $\mathbf{I}$ is the reconstructed image, and $\mathbf{P}'$ and $\mathbf{I}'$ are the corresponding patches and image after the convolutions, respectively. 

\subsection{Reverse Embedding}
First, Reverse Embedding is used to convert the latent vectors from the MHSA into the original shape of the patches. 
In Reverse Embedding, a trainable linear layer is used to restore the dimensionality of the reconstruction, as shown in Algorithm~\ref{alg:vic} Line~2. 
This is used because the dimensions of the embedding $d$ are defined by a hyperparameter and are not necessarily related to the patch size. 
After the embedding of each patch is mapped to a $P_hP_w$-dimensional vector, where $P_h$ is the height of the patch and $P_w$ is the width, the embedding is reshaped to patch $\mathbf{p}_t\in\mathbf{P}$ with the dimensions of the original patch size $(P_h,P_w)$.

\subsection{Reconstruction}
Next, a feature map $\mathbf{I}$ is constructed from the patches. 
Each of the reconstructed patches in $\mathbf{P}$ is concatenated spatially in the position corresponding to the original patch location to form a matrix of the same size as the input image. 
With feature map $\mathbf{I}$, it is possible to treat it as if it were a feature map in a traditional CNN.

\subsection{Convolutional Neural Network (CNN)}

A CNN is used inside the ViC Encoder block instead of the typical MLP. 
The CNN used in the experiments contains one convolutional layer ($C=1$, where $C$ is the number of convolutions). 
However, there is no restriction on the CNN used in the ViC Encoder block. 
The purpose of the CNN is to use information across patches instead of discrete flattened patches like the MLP in ViT. 
Furthermore, the convolutional layers are able to help extract features from the image representation.

\subsection{Patch Embedding}

To continue using the ViC Encoder block as a Transformer block, the output of the CNN needs to be restored to a sequence of token embeddings. 
Thus, we perform the same patch embedding procedure as the input of ViT and the proposed ViC. 
Namely, another trainable linear projection is used to create the patch embeddings $\mathbf{Z}'_\mathrm{patch}$. 

Finally, the classification token $\mathbf{z}_\mathrm{CLS}$ is prepended back to form the full $\mathbf{Z}'$. 
$\mathbf{Z}'$ can now be used with the MLP like a standard Transformer.

\section{Experimental Results}
\label{sec:results}

\subsection{Architecture Settings}

For the experiments, the proposed method uses three of the ViC blocks ($L=3$) shown in Fig.~\ref{fig:encoder}.
Each block uses MHSA with four heads. 
In addition, each block has one convolutional layer with $3\times3$ convolutions at stride 1 and 32 filters each. 
As suggested by Dosovitskiy et al.~\cite{dosovitskiy2020image}, Gaussian Error Linear Units~(GeLU)~\cite{hendrycks2016gaussian} is used as the activation function for all of the trainable layers, including the convolutional layers. 
For the patch encodings, we use $4\times4$ pixel patches and a patch encoding latent space with 256 dimensions. 

The proposed method and all of the comparative evaluations use the same training scheme. 
The networks are trained with batch size 256 for 500 epochs using an Adam optimizer~\cite{kingma2014adam} with an initial learning rate of 0.001 and a weight decay of 0.0005. 
The size of the input, number of input channels, and number of classes is determined by each dataset. 
In addition, we use the pre-defined training and test sets that were determined by the dataset authors.

\subsection{Comparative Evaluations}
In order to evaluate the proposed method, we compare it to other ViT models that incorporate convolutions in some aspect. 
For a fair comparison, the shared hyperparameters of each of the comparison methods were set to match the proposed method. 
Namely, three Transformer blocks with four heads are used. 
In addition, all of the comparative methods use the same $4\times4$ pixel patches and a 256 dimensional linear projection embedding.
They are all trained for the same 500 epochs with Adam optimizer and an initial learning rate of 0.001 and a weight decay of 0.0005. 
All of the networks are trained without pre-training or data augmentation. 
The following comparative evaluations were performed:
\begin{itemize}
    \item \textbf{Vision Transformer (ViT)}. This is the baseline used to demonstrate the usefulness of the proposed method. 
    The ViT evaluation uses all of the same hyperparameters, except for the convolutional layers and reverse embedding layers. 
    \item \textbf{Compact Convolutional Transformer~(CCT)}~\cite{hassani2021escaping}. CCT uses two convolutional layers instead of the traditional embedding layer in ViT. The implementation uses convolutions with kernel size 3 at stride 1 like the proposed method. In addition, CCT uses Sequence Pooling for the MLP head used for classification.
    \item \textbf{Convolution and Self-Attention Network~(CoAtNet)}~\cite{dai2021coatnet}. CoAtNet combines a CNN with ViT by having the lower layers be CNN blocks while the higher layers be Transformer blocks. We use the C-T-T-T version of CoAtNet for comparison as it consists of one convolutional block, one depth-wise convolutional block~\cite{Sandler_2018}, and three Transformer blocks. 
    \item \textbf{Convolutional vision Transformer~(CvT)}~\cite{wu2021cvt}. CvT uses convolutions in two parts of the model. First, convolutions are used in the token embeddings. Second, convolutions are used for the projections for the self-attention layer. Again, for the experiments, the hyperparameters used were set to match the proposed method. 
    \item \textbf{Nested Hierarchical Transformer~(NesT)}~\cite{zhang2022nested}. NesT incorporates a convolution into the aggregation function of hierarchical Transformer blocks. NesT-T is used for the evaluation which includes three hierarchical layers of 8, 4, and 1 Transformer blocks each.
    \item \textbf{Pooling-based Vision Transformer~(PiT)}~\cite{heo2021rethinking}. PiT uses strided convolutions to downsample the patch token embeddings. To match the proposed method, one Transformer block is used between each downsampling, for a total of three Transformer blocks.
\end{itemize}

\begin{table*}[]
    \caption{Average Test Accuracy (\%) of Five Trainings}
    \centering
    \begin{tabular}{l|ccc}
        \toprule
        Model& MNIST & EMNIST & KMNIST \\ \midrule
        ViC (Proposed)& \textbf{99.03} & 87.74 & \textbf{95.86} \\
        ViT& 98.34 & 86.81 & 92.92 \\ \midrule
        CoAtNet& 98.82 & 86.85 & 94.56 \\
        CCT& 98.68 & 87.75 & 94.47 \\
        CvT& 98.72 & \textbf{88.02} & 94.86 \\
        NesT& 98.80 & 85.02 & 94.68 \\
        PiT& 98.77 & 87.39 & 94.91 \\
        \bottomrule
    \end{tabular}
    \label{tab:resultscharacter}
\end{table*}

\subsection{Results on MNIST}
\subsubsection{Dataset}

The Modified National Institute of Standards and Technology (MNIST) database~\cite{Lecun_1998} is a standard benchmark dataset. 
It is made of $28\times28$ pixel, grayscale, isolated handwritten digits. 
There are 10 classes, ``0'' to ``9.''
MNIST has a pre-defined training set of 60,000 images and a test set of 10,000 images.

\subsubsection{Results}

The results of the experiments are shown in Table~\ref{tab:resultscharacter}. 
The results are the mean of training each model five times. 
This is done to increase the reliability of the results. 
In the table, it can be observed that the proposed ViC was able to achieve a higher accuracy than ViT on MNIST. 
In addition, it did remarkably better than the comparison convolutional Transformers. 
ViC had a 99.03\% accuracy, whereas all of the others had less than 99\%. 
It should be noted that the accuracies are lower than some state-of-the-art methods in literature. 
This is because all of the comparisons were evaluated from scratch without the use of techniques such as data augmentation, pre-trained weights, parameter searches, etc.

\subsection{Results on EMNIST}
\subsubsection{Dataset}
Extended MNIST (EMNIST)~\cite{cohen2017emnist} is an extension of MNIST that includes alphabet characters in addition to digits.
For the experiments, we use the ``balanced'' subset. 
The subset includes 112,800 training images and 18,800 test images. 
There are 47 classes, 10 digits and 37 letters. 
The letters include uppercase and lowercase letters in distinct classes, but with certain letters merged due to ambiguity between upper and lowercase. 
The merged letters are ``C,'' ``I,'' ``J,'' ``K,'' ``L,'' ``M,'' ``O,'' ``P,'' ``S,'' ``U,'' ``V,'' ``W,'' ``X,'' ``Y,'' and ``Z.'' 
EMNIST is used because it is similar to MNIST, but is a more difficult problem.

\subsubsection{Results}
In Table~\ref{tab:resultscharacter}, the results for EMNIST are also shown. 
For EMNIST, CvT and CCT had higher accuracies than the proposed method. 
However, the proposed method still outperformed the standard ViT. 
The CvT evaluation performed better than all of the comparison methods.     

\subsection{Results on KMNIST}
\subsubsection{Dataset}

\begin{figure}
\centering
\includegraphics[width=1\linewidth]{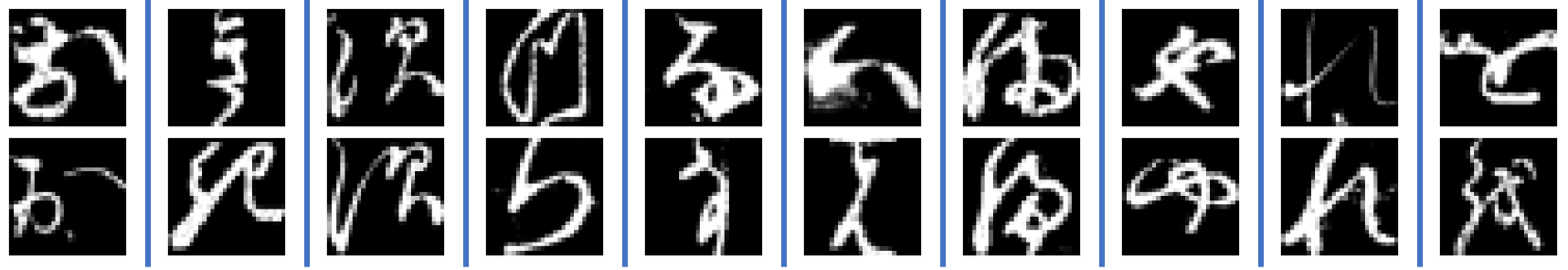}%
\caption{Two examples from each class of KMNIST}
\label{fig:kmnist}
\end{figure}

The final character recognition dataset used is Kuzushiji-MNIST (KMNIST) dataset~\cite{clanuwat2018deep}.
KMNIST consists of 10 classes of kurzushiji, or Japanese cursive. 
The 10 classes represent 10 of the classical Japanese Hiragana characters.
An example from each class is shown in Fig.~\ref{fig:kmnist}.
Similar to MNIST, the images are $28\times28$ grayscale isolated characters. 
There are 60,000 training images and 10,000 test images.
KMNIST is also used as a baseline with a more difficult character recognition task.

\subsubsection{Results}

KMNIST had the largest discrepancy between the proposed method and ViT. 
ViT only had an average accuracy of 92.92\%, whereas all of Transformers that include convolutions had 94\% or higher. 
The proposed method performed the best at 95.86\% accuracy.
One possible explanation for the increase in accuracy, especially over the original ViT, is that there are more variations of characters within classes of KMNIST. 
In Fig.~\ref{fig:kmnist}, for some characters, there are large deformations. 
To some extent, adding convolutions provides some translation invarience~\cite{Lecun_1998}.

\begin{figure}
    \centering
    \subfloat[MNIST]{
    \label{CNN_layer.sub.a}
    \includegraphics[width=0.3\textwidth]{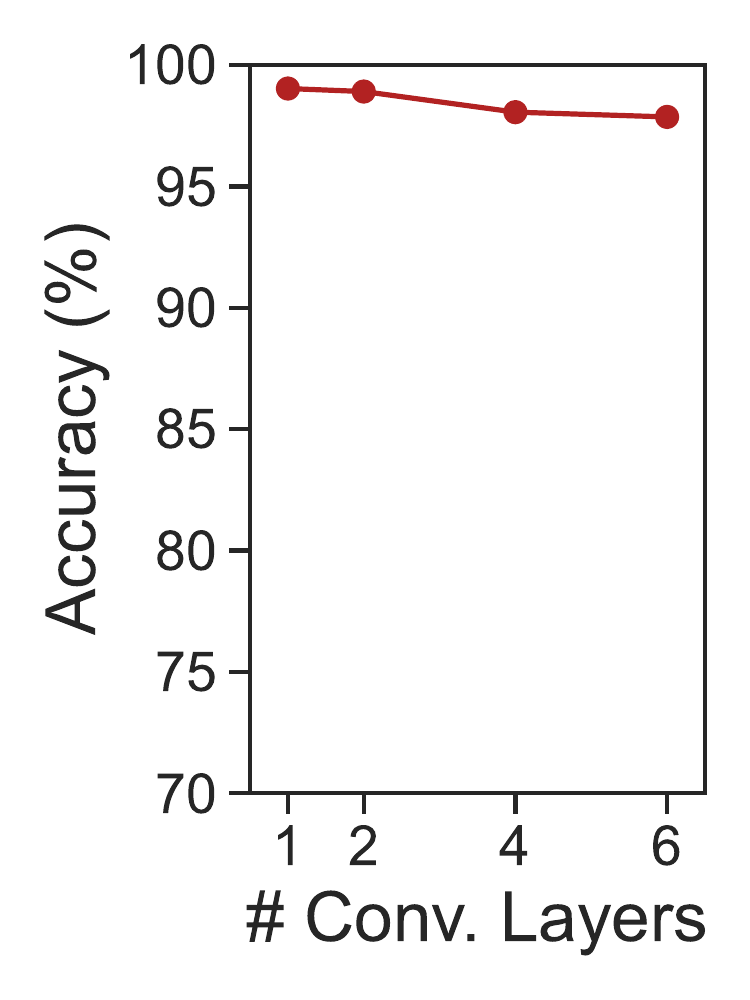}
    }
    \subfloat[EMNIST]{
    \label{CNN_layer.sub.b}
    \includegraphics[width=0.3\textwidth]{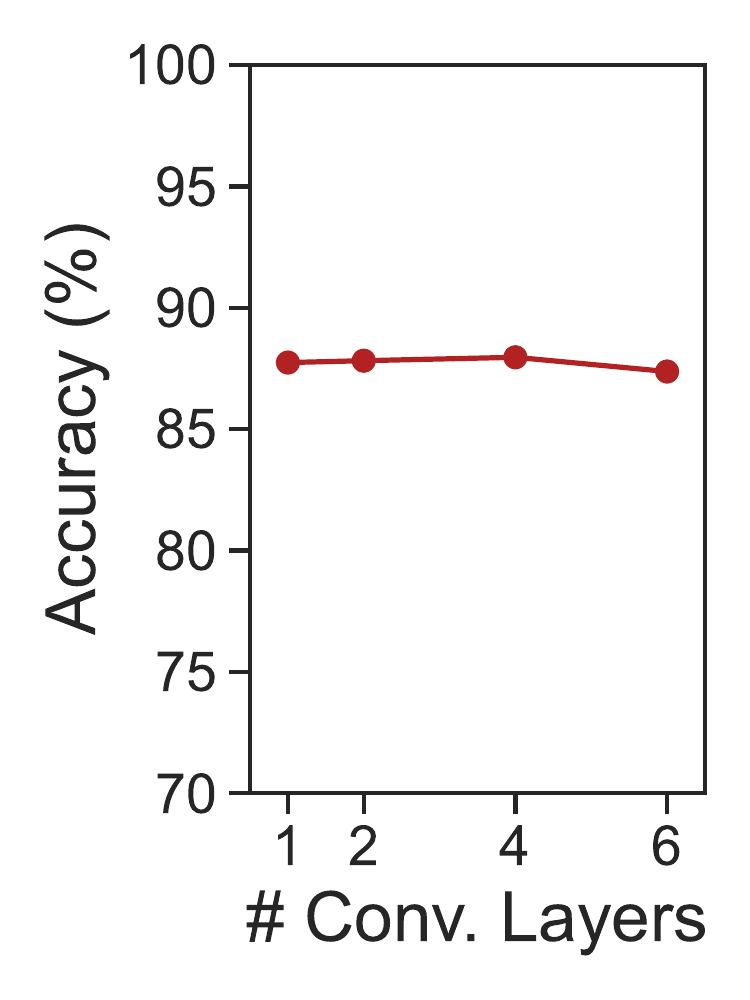}
    }
   \subfloat[KMNIST]{
    \label{CNN_layer.sub.c}
    \includegraphics[width=0.3\textwidth]{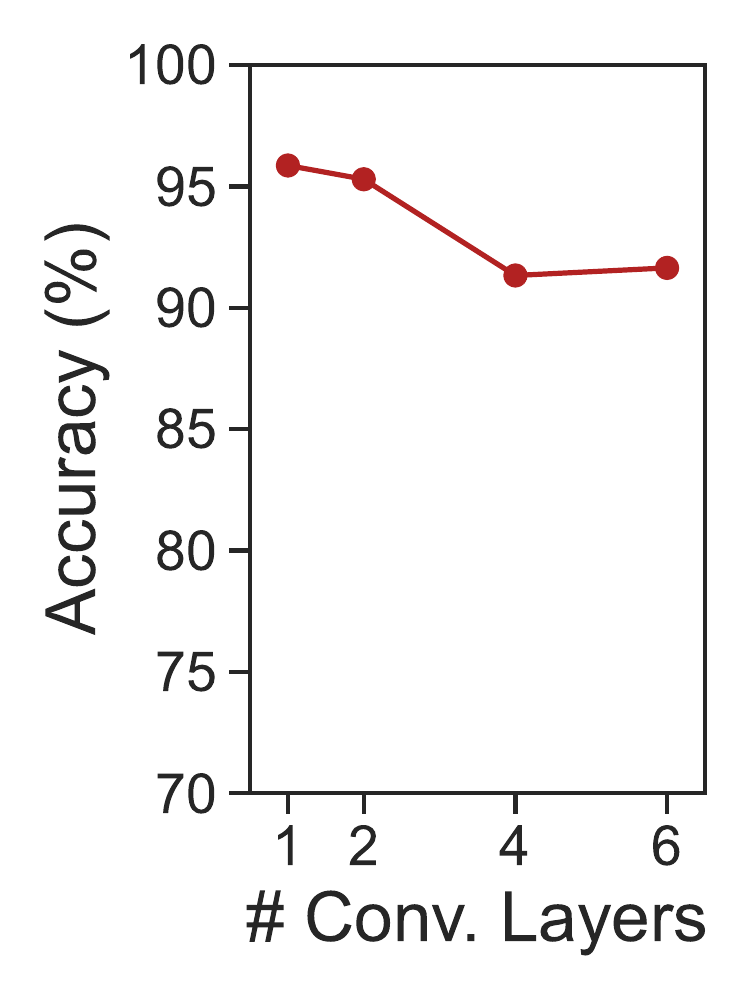}
    }
\caption{The test accuracy comparing the depth of the CNN. These are the average of 5 trainings.}
\label{fig:CNN_layer}
\end{figure}

\section{Ablation Study}

In order to see the effects of the convolutional layers, we perform an ablation study that examines the change in performance while varying the number of convolutional layers in the ViC layers. 
The experimental setup is the same as described in Section \ref{sec:results}.
The results of the experiments are shown in Fig.~\ref{fig:CNN_layer}.
On MNIST and KMNIST, adding additional convolutional layers decreased the accuracy. 
EMNIST had a slight increase in accuracy at two and four layers, but it decreased again at six. 
Thus, in general, increasing the number of layers further would not provide additional benefit.

\section{Application to General Object Recognition}

The proposed method is not limited to only character recognition. 
It can be extended to any image recognition task. 
Thus, we demonstrate the ability to use the proposed method on general object recognition.

\subsection{Datasets}
In order to demonstrate the use of the proposed method on other tasks, we use four datasets. 
The first two datasets are the CIFAR10 and CIFAR100 datasets~\cite{krizhevsky2009learning}.
They are natural scene object datasets with 10 classes and 100 classes, respectively. 
The next dataset is the STL10 dataset~\cite{coates2011analysis}. 
The STL10 dataset is similar to CIFAR10, but contains larger images of $96\times96$ pixels.
The final dataset is the FashionMNIST dataset~\cite{xiao2017fashion}. 
This dataset contains images in a similar style to MNIST, except instead of digits, it is made of articles of clothing.

\subsection{Results}

\begin{table*}[]
    \caption{General Image Average Test Accuracy (\%) of Five Trainings}
    \centering
    \begin{tabular}{l|cccc}
        \toprule
        Model& CIFAR10 & CIFAR100 & FashionMNIST & STL10\\ \midrule
        ViC (Proposed)& \textbf{80.23} & 53.14 & 90.02 & \textbf{58.21}\\
        ViT& 76.60 & \textbf{55.58} & 87.86 & 57.58 \\ \midrule
        CoAtNet & 63.44 & 32.83 & 88.30 & 53.39 \\
        CCT& 77.89 & 50.26 & 89.51 & 51.90 \\
        CvT& 77.99 & 46.33 & 89.18 & 46.94\\
        NesT& 67.96 & 38.10 & 88.34& 51.55\\
        PiT& 69.80 & 41.51 & \textbf{90.19} & 51.55 \\
        \bottomrule
    \end{tabular}
    \label{tab:results}
\end{table*}

The results are shown in Table~\ref{tab:results}. 
Similar to character recognition, the proposed method performed better overall when compared to ViT and convolutional Transformers. 
In every case except CIFAR100, the proposed ViC did better than ViT. 
However, PiT had a higher accuracy than the proposed method on FashionMNIST. 

\section{Conclusion}

In this paper, we presented a new neural network model that combines ViT with a CNN called a Vision Conformer~(ViC). 
Unlike CNNs, ViT has much less image-specific inductive bias~\cite{dosovitskiy2020image}. 
This is because the original transformer was designed for discrete sequences. 
Therefore, in order to add consideration for image structures and local pixel relationships, we introduced convolutions into ViT. 

Specifically, we proposed replacing the MLP inside the Transformer block of ViT with a CNN. 
In order to be able to do this, the internal representation should be an image-like feature map. 
Thus, we propose a process of reverse embedding and reconstruction. 
In the reverse embedding, we perform the opposite operation of the patch embedding process. 
Namely, the latent token embeddings from MHSA are embedded into a flattened patch-sized space using a trainable linear projection. 
Next, the embeddings are reshaped into patches and reconstructed into a feature map using spatial concatenation. 

We evaluated the proposed method on character recognition. 
Namely, we demonstrated that the proposed method was effective on the MNIST, EMNIST, and KMNIST datasets. 
% the CIFAR10, CIFAR100, Fashion MNIST, and STL10 datasets. 
Through the experiments, we demonstrate that the proposed method can improve the classification ability of ViT. 
In addition, we showed that similar results were found on general image recognition tasks.
In the future, work will be done exploring the complexity and improving upon the CNN within the proposed ViC.

\bibliographystyle{splncs04}
\bibliography{vic}

\end{document}